\let\accentvec\vec
\documentclass[runningheads]{packages/llncs}

\let\vec\accentvec

\usepackage{graphicx}
\usepackage{packages/comment}
\usepackage{amsmath,amssymb} 
\usepackage{color}

\usepackage[colorlinks]{hyperref}
\usepackage{multirow} 
\usepackage{array}
\newcolumntype{C}[1]{>{\centering\arraybackslash}p{#1}}

\usepackage{etoolbox}
\usepackage{enumitem}

\usepackage{bm}
\usepackage[ruled, vlined]{algorithm2e}

\usepackage{makecell}

\usepackage{bbding}

\usepackage{tikz}
\usepackage{pgfplots}

\usepackage{wrapfig}

\usepackage{sidecap}

\usepackage{bbm} 

\usepackage{listings}

\usepackage[caption = false]{subfig}
\newcommand{\Noindent}{\noindent}
\newcommand{\partitle}[1]{\Noindent\textbf{#1. }}
\newcommand{\parclaim}[1]{\Noindent\textbf{#1: }}

\newcommand{\secbest}[1]{\underline{#1}}

\newcommand{\xmark}{\ding{55}}%
\newcommand{\cmark}{\ding{51}}

\renewcommand{\arraystretch}{1}

\def\ours{AR-Net }
\def\ourswithoutspace{AR-Net}
\def\ECCVSubNumber{152}  
\title{\ourswithoutspace: Adaptive Frame Resolution for Efficient Action Recognition}

\newtoggle{CameraReady}
\settoggle{CameraReady}{true} 

\newtoggle{Arxiv}
\settoggle{Arxiv}{true}

\DeclareMathOperator{\E}{\mathbb{E}}

\renewcommand\figurename{Figure }

\pgfdeclareplotmark{lwjx}{ 
\draw  [color=green!60!black ,draw opacity=0 ][fill=green!60!black  ,fill opacity=1 ]
(0.0000\pgfplotmarksize,1.8542\pgfplotmarksize) --
(0.4482\pgfplotmarksize,0.5724\pgfplotmarksize) --
(1.9000\pgfplotmarksize,0.5729\pgfplotmarksize) --
(0.7250\pgfplotmarksize,-0.2187\pgfplotmarksize) --
(1.1744\pgfplotmarksize,-1.5000\pgfplotmarksize) --
(0.0000\pgfplotmarksize,-0.7073\pgfplotmarksize) --
(-1.1744\pgfplotmarksize,-1.5000\pgfplotmarksize) --
(-0.7250\pgfplotmarksize,-0.2187\pgfplotmarksize) --
(-1.9000\pgfplotmarksize,0.5729\pgfplotmarksize) --
(-0.4482\pgfplotmarksize,0.5724\pgfplotmarksize) --
  cycle;
}

\pgfdeclareplotmark{wjx}{ 
\draw  [color={rgb, 255:red, 255; green, 0; blue, 0 }  ,draw opacity=0 ][fill={rgb, 255:red, 255; green, 0; blue, 0 }  ,fill opacity=1 ]
(0.0000\pgfplotmarksize,1.8542\pgfplotmarksize) --
(0.4482\pgfplotmarksize,0.5724\pgfplotmarksize) --
(1.9000\pgfplotmarksize,0.5729\pgfplotmarksize) --
(0.7250\pgfplotmarksize,-0.2187\pgfplotmarksize) --
(1.1744\pgfplotmarksize,-1.5000\pgfplotmarksize) --
(0.0000\pgfplotmarksize,-0.7073\pgfplotmarksize) --
(-1.1744\pgfplotmarksize,-1.5000\pgfplotmarksize) --
(-0.7250\pgfplotmarksize,-0.2187\pgfplotmarksize) --
(-1.9000\pgfplotmarksize,0.5729\pgfplotmarksize) --
(-0.4482\pgfplotmarksize,0.5724\pgfplotmarksize) --
  cycle;
}

\def\depict{Videos are uniformly sampled in 8 frames. The first row in each example is the original video input, and the second row represents the resolutions or skipping decisions that AR-NET chooses. We shows ground truth labels and define ``difficulties" (Easy, Medium and Hard) based on their computation budgets. AR-Net can save the computation greatly for examples which contain clear appearance or actions with less motion. Best viewed in color.}
\nottoggle{CameraReady}{ 
\usepackage{packages/ruler}
\usepackage[width=122mm,left=12mm,paperwidth=146mm,height=193mm,top=12mm,paperheight=217mm]{geometry}
}{}

\nottoggle{Arxiv}{}{
\usepackage[width=122mm,left=12mm,paperwidth=146mm,height=193mm,top=12mm,paperheight=217mm]{geometry}
}

\usepackage{csquotes}
\usepackage{pifont}

\pgfplotsset{compat=1.14}
\begin{document}
\pagestyle{headings}
\mainmatter

\nottoggle{CameraReady}{
\titlerunning{ECCV-20 submission ID \ECCVSubNumber} 
\authorrunning{ECCV-20 submission ID \ECCVSubNumber} 
\author{Anonymous ECCV submission}
\institute{Paper ID \ECCVSubNumber}}{}
\iftoggle{CameraReady}{
\titlerunning{Adaptive Resolution for Efficient Action Recognition}
\author{
Yue Meng\inst{1} \and
Chung-Ching Lin\inst{1} \and
Rameswar Panda \inst{1} \and
Prasanna Sattigeri \inst{1} \and
Leonid Karlinsky \inst{1} \and
Aude Oliva \inst{1,3} \and
Kate Saenko \inst{1,2} \and
Rogerio Feris \inst{1}
}

\authorrunning{Y. Meng et al.}

\institute{MIT-IBM Watson AI Lab, IBM Research \and Boston University\and Massachusetts Institute of Technology}
}{}

\maketitle
\begin{abstract}

Action recognition is an open and challenging problem in computer vision. While current state-of-the-art models offer excellent recognition results, their computational expense limits their impact for many real-world applications. In this paper, we propose a novel approach, called \ours (Adaptive Resolution Network), that selects on-the-fly the optimal resolution for each frame conditioned on the input for efficient action recognition in long untrimmed videos.
Specifically, given a video frame, a policy network is used to decide what input resolution should be used for processing by the action recognition model, with the goal of improving both accuracy and efficiency. 
We efficiently train the policy network jointly with the recognition model using standard back-propagation. Extensive experiments on several challenging action recognition benchmark datasets well demonstrate the efficacy of our proposed approach over state-of-the-art methods. The project page can be found at \url{https://mengyuest.github.io/AR-Net}

\end{abstract}
\section{Introduction}
\begin{figure}[!htbp]
    \centering
    \includegraphics[width=.99\linewidth]{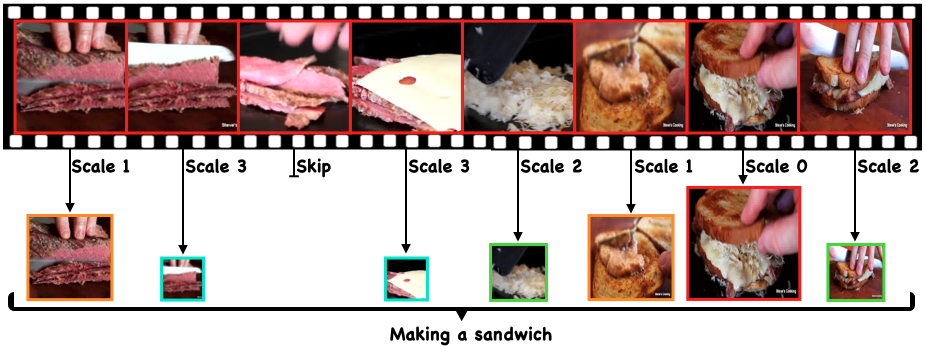} 
    \caption{\textbf{A conceptual overview of our approach.} Rather than processing all the frames at the same resolution, our approach learns a policy to select the optimal resolution (or skip) per frame, that is needed to correctly recognize an action in a given video. As can be seen from the figure, the seventh frame is the most useful frame for recognition, therefore could be processed only with the highest resolution, while the rest of the frames could be processed at lower resolutions or even skipped without losing any accuracy. Best viewed in color.}
    \label{fig:arch_graph}
\end{figure}

Action recognition has attracted intense attention in recent years. Much progress has been made in developing a variety of ways to recognize complex actions, by either applying 2D-CNNs with additional temporal modeling~\cite{karpathy2014large,wang2016temporal,fan2019more} or 3D-CNNs that model the space and time dimensions jointly~\cite{tran2015learning,carreira2017quo,hara2018can}. Despite impressive results on commonly used benchmark datasets, the accuracy obtained by most of these models usually grows proportionally with their complexity and computational cost. This poses an issue for deploying these models in many resource-limited applications such as autonomous vehicles and mobile platforms. 

Motivated by these applications, extensive studies have been recently conducted for designing compact architectures~\cite{piergiovanni2019tiny,howard2017mobilenets,iandola2016squeezenet,zhang2018shufflenet,araujo2018training} or compressing models~\cite{dong2017more,wen2017coordinating,chen2015compressing,li2016pruning}. However, most of the existing methods process all the frames in a given video at the same resolution. In particular, orthogonal to the design of compact models, the computational cost of a CNN model also has much to do with the input frame size. To illustrate this, let us consider the video in Figure~\ref{fig:arch_graph}, represented by eight uniformly sampled frames. We ask, \textit{Do all the frames need to be processed at the highest resolution (e.g., 224$\times$224) to recognize the action as \enquote{Making a sandwich} in this video?} The answer is clear: No, the seventh frame is the most useful frame for recognition, therefore we could process only this frame at the highest resolution, while the rest of the frames could be processed at lower resolutions or even skipped (i.e., resolution set to zero) without losing any accuracy, resulting in large computational savings compared to processing all the frames with the same 224$\times$224 resolution. Thus, in contrast to the commonly used one-size-fits-all scheme, we would like these decisions to be made individually per input frame, leading to different amounts of computation for different videos. Based on this intuition, we present a new perspective for efficient action recognition by adaptively selecting input resolutions, on a per frame basis, for recognizing complex actions.

In this paper, we propose \ourswithoutspace, a novel and differentiable approach to learn a decision policy that selects optimal frame resolution conditioned on inputs for efficient action recognition. The policy is sampled from a discrete distribution parameterized by the output of a lightweight neural network (referred to as the policy network), which decides on-the-fly what input resolution should be used on a per frame basis. As these decision functions are discrete and non-differentiable, we rely on a recent Gumbel Softmax sampling approach~\cite{jang2016categorical} to learn the policy jointly with the network parameters through standard back-propagation, without resorting to complex reinforcement learning as in~\cite{wu2019adaframe,fan2018watching,yeung2016end}.
We design the loss to achieve both competitive performance and resource efficiency required for action recognition. We demonstrate that adaptively selecting the frame resolution by a lightweight policy network yields not only significant savings in FLOPS (e.g., about 45\% less computation over a state-of-the-art method~\cite{wu2019liteeval} on ActivityNet-v1.3 dataset~\cite{caba2015activitynet}), but also consistent improvement in action recognition accuracy. 

The main contributions of our work are as follows:
\begin{itemize}
    \item We propose a novel approach that automatically determines what resolutions to use per target instance for efficient action recognition. 
    \item We train the policy network jointly with the recognition models using backpropagation through Gumbel Softmax sampling, making it highly efficient.
    \item We conduct extensive experiments on three benchmark datasets (ActivityNet-V1.3~\cite{caba2015activitynet}, FCVID~\cite{jiang2017exploiting} and Mini-Kinetics~\cite{carreira2017quo}) to demonstrate the superiority of our proposed approach over state-of-the-art methods.
\end{itemize}
\section{Related Works}

\textbf{Efficient Action Recognition.} Action recognition has made rapid progress with the introduction of a number of large-scale datasets such as Kinetics~\cite{carreira2017quo} and Moments-In-Time~\cite{monfort2019moments,monfort2019multi}. 
Early methods have studied action recognition using shallow classification models such as SVM on top of local visual features extracted from a video~\cite{laptev2008learning,wang2011action}.
In the context of deep neural networks, it is typically performed by either 2D-CNNs~\cite{karpathy2014large,simonyan2014two,cheron2015p,feichtenhofer2017spatiotemporal,gkioxari2015finding} or 3D-CNNs~\cite{tran2015learning,carreira2017quo,hara2018can}. A straightforward but popular approach is the use of 2D-CNNs to extract frame-level features and then model the temporal causality across frames using different aggregation modules such as temporal averaging in TSN~\cite{wang2016temporal}, a bag of features scheme in TRN~\cite{zhou2018temporal}, channel shifting in TSM~\cite{lin2019tsm}, depthwise convolutions in TAM~\cite{fan2019more}, non-local neural networks~\cite{wang2018non}, and LSTMs~\cite{donahue2015long}. Many variants of 3D-CNNs such as C3D~\cite{tran2015learning}, I3D~\cite{carreira2017quo} and ResNet3D~\cite{hara2018can}, that use 3D convolutions to model space and time jointly, have also been introduced for action recognition. 

While extensive studies have been conducted in the last few years, limited efforts have been made towards {\em efficient} action recognition~\cite{wu2019adaframe,wu2019liteeval,gao2019listen}. Specifically, methods for efficient recognition focus on either designing new lightweight architectures (e.g., R(2+1)D~\cite{tran2018closer}, Tiny Video Networks~\cite{piergiovanni2019tiny}, channel-separated CNNs~\cite{tran2019video}) or selecting salient frames/clips conditioned on the input~\cite{yeung2016end,wu2019adaframe,korbar2019scsampler,gao2019listen}. Our approach is most related to the latter which focuses on adaptive data sampling and is agnostic to the network architecture used for recognizing actions. Representative methods typically use Reinforcement Learning (RL) where an agent~\cite{wu2019adaframe,fan2018watching,yeung2016end} or multiple agents~\cite{wu2019multi} are trained with policy gradient methods to select relevant video frames, without deciding frame resolution as in our approach. More recently, audio has also been used as an efficient way to select salient frames for action recognition~\cite{korbar2019scsampler,gao2019listen}. Unlike existing works, our framework requires neither complex RL policy gradients nor additional modalities such as audio. LiteEval~\cite{wu2019liteeval} proposes a coarse-to-fine framework for resource efficient action recognition that uses a binary gate for selecting either coarse or fine features. In contrast, we address both the selection of optimal frame resolutions and skipping in an unified framework and jointly learn the selection and recognition mechanisms in a fully differentiable manner. Moreover, unlike binary sequential decision being made at every step in LiteEval, our proposed approach has the flexibility in deciding multiple actions in a single step and also the scalability towards long untrimmed videos via multi-step skipping operations. We include a comprehensive comparison to LiteEval in our experiments.

\noindent\textbf{Adaptive Computation.} Many adaptive computation methods have been recently proposed with the goal of improving computational efficiency~\cite{bengio2015conditional,bengio2013estimating,veit2018convolutional,wang2018skipnet,graves2016adaptive}.
Several works have been proposed that add decision branches to different layers of CNNs to learn whether to exit the network for faster inference~\cite{figurnov2017spatially,mcgill2017deciding}.
BlockDrop~\cite{wu2018blockdrop} effectively reduces the inference time by learning to dynamically select which
layers to execute per sample during inference. 
Adaptive computation time for recurrent neural networks is also presented in~\cite{graves2016adaptive}.
SpotTune~\cite{guo2019spottune} learns to adaptively route information through finetuned or pre-trained layers. Reinforcement learning has been used to adaptively select different regions for fast object detection in large images~\cite{najibi2019autofocus,gao2018dynamic}. While our approach is inspired by these methods, in this paper, we focus on adaptive computation in videos, where our goal is to adaptively select optimal frame resolutions for efficient action recognition.

\noindent\textbf{Multi-Resolution Processing.}  Multi-resolution feature representations have a long history in computer vision. Traditional methods include image pyramids~\cite{adelson1984pyramid}, scale-space representations~\cite{perona1990scale}, and coarse-to-fine approaches~\cite{pedersoli2015coarse}. More recently, in the context of deep learning, multi-scale feature representations have been used for detection and recognition of objects at multiple scales \cite{cai2016unified,lin2017feature}, as well as to speed up deep neural networks \cite{lin2017feature,chen2018big}. Very few approaches have explored multi-scale recognition for efficient video understanding. A two-branch network that fuses the information of high-resolution and low-resolution video frames is proposed in~\cite{karpathy2014large}. bLVNet-TAM \cite{fan2019more} also uses a two-branch multi-resolution architecture based on the Big-Little Net model \cite{chen2018big}, while learning long-term temporal dependencies across frames. SlowFast Networks~\cite{feichtenhofer2019slowfast} rely on a similar two-branch model, but each branch encodes different frame rates (i.e., different temporal resolutions), as opposed to frames with different spatial resolutions. Unlike these methods, rather than processing video frames at multiple resolutions with specialized network branches, our approach determines optimal resolution for each frame, with the goal of improving accuracy and efficiency.
\section{Proposed Method}

Given a video dataset $\mathcal{D}=\{(V_i,y_i)\}_{i=1}^N$, where each video $V_i$ contains frames with spatial resolution $3\times H_0\times W_0$ and is labelled from the predefined classes: $y_i\in\mathbb{Y}= \{0,1,...,C-1\}$, our goal is to create an adaptive selection strategy that decides, per input frame, which resolution to use for processing by the classifier $\mathcal{F}: \mathbb{V}\to \mathbb{Y}$ with the goal of improving accuracy and efficiency. To this end, we first present an overview of our approach in Section~\ref{section:overview}. Then, we show how we learn the decision policy using Gumbel Softmax sampling in Section~\ref{section:learning}. Finally, we discuss the loss functions used for learning the decision policy in Section~\ref{section:loss}.

\begin{figure*} [t]
	\centering
	\begin{tabular}{c}
		\includegraphics[scale=0.288]{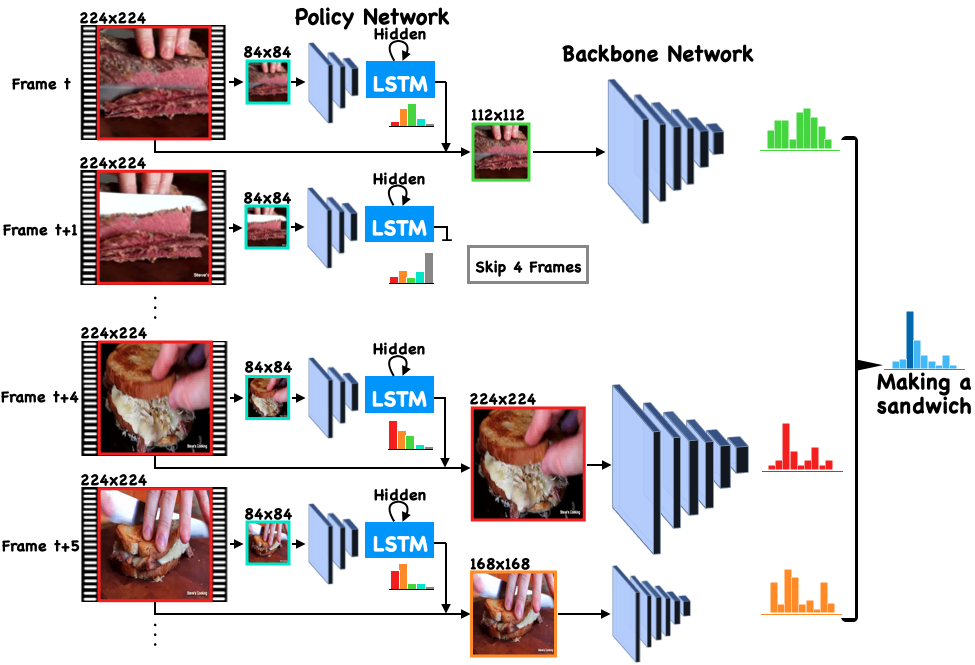}
	\end{tabular} 
	\caption
	{\textbf{Illustration of our approach}. \ours consists of a policy network and different backbone networks corresponding to different resolutions. The policy network decides what resolution (or skip) to use on a per frame basis to achieve accuracy and efficiency. In training, policies are sampled from a Gumbel Softmax distribution, which allows to optimize the policy network via backpropagation. During inference, input frames are first fed into the policy network to decide the proper resolutions, then the rescaled frames are routed to corresponding backbones to generate predictions. Finally the network averages all the predictions for action classification. Best viewed in color.}
	 \label{fig:arch}
\end{figure*}

\subsection{Approach Overview}
\label{section:overview}
Figure \ref{fig:arch} illustrates an overview of our approach, which consists of a policy network and backbone networks for classifying actions. The policy network contains a lightweight feature extractor and an LSTM module that decides what resolutions (or skipping) to use per input frame, for efficient action recognition. Inspired by the compound scaling method~\cite{tan2019efficientnet}, we adopt different network sizes to handle different resolutions, as a frame with a higher resolution should be processed by a heavier network because of its capability to handle the detailed visual information and vice versa.Furthermore, it is often unnecessary and inefficient to process every frame in a video due to large redundancy coming from static scenes or the frame quality being very low (blur, low-light condition, etc). Thus, we design a skipping mechanism in addition to the adaptive selection of frame resolutions in an unified framework to skip frames (i.e., resolution set to zero) whenever necessary to further improve the efficiency in action recognition.  

During training, the policy network is jointly trained with the recognition models using Gumbel Softmax sampling, as we will describe next. At test time, an input frame is first fed into a policy network, whose output decides the proper resolutions, and then the resized frames are routed to the corresponding models to generate the predictions. Finally, the network averages all the predictions as the action classification result. Note that the additional computational cost is incurred by resizing operations and the policy network, which are negligible in comparison to the original recognition models (the policy network is designed to be very lightweight, e.g., MobileNetv2 in our case).

\subsection{Learning the Adaptive Resolution Policy}
\label{section:learning}
\partitle{Adaptive Resolution} \ours adaptively chooses different frame scales to achieve efficiency. Denote a sequence of resolutions in descending order as $\{s_i\}_{i=0}^{L-1}$, where $s_0=(H_0,W_0)$ stands for the original (also the highest) frame resolution, and $s_{L-1}=(H_{L-1},W_{L-1})$ is the lowest resolution. The frame at time $t$ in the $l^{\text{th}}$ scale (resolution $s_l=(H_l,W_l)$) is denoted as $I_t^l$. We consider skipping frames as a special case ``choosing resolutions $s_{\infty}$". We define the skippings sequence (ascending order) as $\{F_i\}_{i=0}^{M-1}$, where the $i^{\text{th}}$ operation means to skip the current frame and the following $(F_i-1)$ frames from predictions. The choices for resolutions and skippings formulate our action space $\Omega$.

\noindent\textbf{Policy Network.} The policy network contains a lightweight feature extractor $\Phi(\cdot; \theta_\Phi)$ and an LSTM module. At time step $t<T$ we resize the frame $I_t$ to the lowest resolution $I_t^{L-1}$ (for efficiency) and send it to the feature extractor,
\begin{equation}
f_t=\Phi(I_t^{L-1};\theta_\Phi)
\end{equation}
where $f_t$ is a feature vector and $\theta_\Phi$ denotes learnable parameters (we use $\theta_{\text{name}}$ for the learnable parameters in the rest of this section). The LSTM updates hidden state $h_t$ and outputs $o_t$ using the extracted feature and previous states,
\begin{equation}
    h_t, o_t = \text{LSTM}(f_t, h_{t-1}, o_{t-1};\theta_{LSTM})
\end{equation}
Given the hidden state, the policy network estimates the policy distribution and samples the action $a_t\in\Omega= \{0,1,...L+M-1\}$ via the Gumbel Softmax operation (will be discussed in the next section),
\begin{equation}
    a_t\sim\text{GUMBEL}(h_t,\theta_G)
    \label{eq:pi}
\end{equation}
If $a_t<L$, we resize the frame to spatial resolution $3\times H_{a_t}\times W_{a_t}$ and forward it to the corresponding backbone network $\Psi_{a_t}(\cdot; \theta_{\Psi_{a_t}})$ to get a frame-level prediction, 
\begin{equation}
   y_t^{a_t}=\Psi_{a_t}(I_t^{a_t}; \theta_{\Psi_{a_t}})
\end{equation}
where $I_t^{a_t}\in \mathbb{R}^{3\times H_{a_t}\times W_{a_t}}$ is the resized frame  and $y_t^{a_t}\in \mathbb{R}^{C}$ is the prediction. Finally, all the frame-level predictions are averaged to generate the video-level prediction $y$ for the given video $V$. 

When the action $a_t>=L$, the backbone networks will skip the current frame for prediction, and the following $(F_{a_t-L}-1)$ frames will be skipped by the policy network. Moreover, to save the computation, we share the policy network for generating both policy and predictions for the lowest resolution, i.e., $\Psi_{L-1}=\Phi$\footnote{The notation here is for brevity. Actually, the output for $\Phi$ is a feature vector, whereas the output for $\Psi_{L-1}$ is a prediction. In implementation, we use a fully connected layer after the feature vector to get the prediction}. 

\partitle{Training using Gumbel Softmax Sampling}
\ours makes decisions about which resolutions (or skipping) to use per training example. 
However, the fact that the decision policy is discrete makes the network non-differentiable and therefore difficult to optimize via backpropagation.
One common practice is to use a score function estimator (e.g.,  REINFORCE \cite{williams1992simple,glynn1990likelihood}) to avoid back-propagating through the discrete samples. However, due to the undesirable fact that the variance of the score function estimator scales linearly with the discrete variable dimension (even when a variance reduction method is adopted), it is slow to converge in many applications \cite{wu2019liteeval,jang2016categorical}. As an alternative, in this paper, we adopt Gumbel-Softmax Sampling~\cite{jang2016categorical} to resolve this non-differentiability and enable direct optimization of the discrete policy in an efficient way.

The Gumbel Softmax trick~\cite{jang2016categorical} is a simple and effective way to substitute the original non-differentiable sample from a discrete distribution with a differentiable sample from a corresponding Gumbel-Softmax distribution. Specifically, at each time step $t$, we first generate the logits  $z\in\mathbb{R}^{L+M-1}$ from hidden states $h_t$ by a fully-connected layer $z=\text{FC}(h_t,\theta_{FC})$. Then we use Softmax to generate a categorical distribution $\bm{\pi}_t$, 
\begin{equation}
    \bm{\pi}_t=\left\{p_i\left|\, p_i=\frac{\exp(z_i)}{\sum_{j=0}^{L+M-1}\exp(z_j)}\right.\right\}
\end{equation}
With the Gumbel-Max trick \cite{jang2016categorical}, the discrete samples from a categorical distribution are drawn as follows:
\begin{equation} 
\label{eq:gumbelmax}
    \hat{p} = \operatorname*{arg\,max}_i (\log p_i+G_i), 
\end{equation}
where $G_i=-\log(-\log U_i)$ is a standard Gumbel distribution with $U_i$ sampled from a uniform i.i.d  distribution $Unif(0,1)$.
Due to the non-differentiable property of $\operatorname*{arg\,max}$ operation in Equation \ref{eq:gumbelmax}, the Gumbel Softmax distribution \cite{jang2016categorical} is thus used as a continuous relaxation to $\operatorname*{arg\,max}$. Accordingly, sampling from a Gumbel Softmax distribution allows us to backpropagate from the discrete samples to the policy network. Let $\hat{P}$ be a one hot vector $[\hat{P}_0, ..., \hat{P}_{L+M-1}]$: 
\begin{equation}
    \hat{P}_i=\begin{cases}
    1,& \text{if } i= \hat{p}\\
    0,              & \text{otherwise}
    \end{cases}
\end{equation}
The one-hot coding of vector $\hat{P}$ is relaxed to a real-valued vector $P$ using softmax: 
\begin{equation}
\label{eq:one}
    P_i=\frac{\exp((\log p_i+G_i)/\tau)}{\sum_{j=0}^{L+M-1} \exp((\log p_j+G_j)/\tau)},\ \ \ \ \ i \in [0, ..., L+M-1]
\end{equation}
where $\tau$ is a temperature parameter, which controls the `smoothness` of the distribution $P$, as  $\lim\limits_{\tau\to +\infty} P$ converges to a uniform distribution and $\lim\limits_{\tau\to 0} P$ becomes a one-hot vector. We set $\tau$ = 5 as the initial value and gradually anneal it down to 0 during the training, as in~\cite{guo2019spottune}. 

To summarize, during the forward pass, we sample the decision policy using Equation~\ref{eq:gumbelmax} (this is equivalent to the process mentioned in Equation \ref{eq:pi} and $\theta_{FC}=\theta_G$) and during the backward pass, we approximate the gradient of the discrete samples by computing the gradient of
the continuous softmax relaxation in Equation~\ref{eq:one}. 

\subsection{Loss Functions}
\label{section:loss}
During training, we use the standard cross-entropy loss to measure the classification quality as:
\begin{equation}
    \mathcal{L}_{acc}=\E_{(V,y)\sim \mathcal{D}_{train}}\left[-y\log(\mathcal{F}(V; \Theta))\right]
\end{equation}
where $\Theta=\{\theta_{\Phi},\theta_{LSTM},\theta_G,\theta_{\Psi_0},...,\theta_{\Psi_{L-2}}\}$ and $(V,y)$ is the training video sample with associated one-hot encoded label vector. The above loss only optimizes for accuracy without taking efficiency into account. To address computational efficiency, we compute the GFLOPS for each individual module (and specific resolution of frames) offline and formulate a lookup table. We estimate the overall runtime GFLOPS for our network based on the offline lookup table $\text{GFLOPS}_\mathcal{F}:\Omega\to\mathbb{R}^+$and online policy $a_{V,t}$ for each training video $(V,y)\sim\mathcal{D}_{train}$. We use the GFLOPS per frame as a loss term to punish for high-computation operations,
\begin{equation}
    \mathcal{L}_{flops}=\E_{(V,y)\sim \mathcal{D}_{train}}\left[\frac{1}{T}\sum\limits_{t=0}^{T-1} \text{FLOPS}_{\mathcal{F}}(a_{V,t})\right]
\end{equation}

Furthermore, to encourage the policy learning to choose more frames for skipping, we add an additional regularization term to enforce a balanced policy usage,
\begin{equation}
    \mathcal{L}_{uni}=\sum\limits_{i=0}^{L+M-1}\left(\mathbb{E}_{(V,y)\sim \mathcal{D}_{train}}\left[\frac{1}{T}\sum\limits_{t=0}^{T-1}\mathbbm{1}(a_{V,t}=i)\right]-\frac{1}{L+M}\right)^2
\end{equation}
where $\mathbbm{1}(\cdot)$ is the indicator function. Here $\mathbb{E}_{(V,y)\sim \mathcal{D}_{train}}\left[\frac{1}{T}\sum\limits_{t=0}^{T-1}\mathbbm{1}(a_{V,t}=i)\right]$ represents the frequency of action $i$ being made through the dataset. Intuitively, this loss function term drives the network to balance the policy usage in order to obtain a high entropy for the action distribution.
To sum up, our final loss function for the training becomes:
\begin{equation}
    \mathcal{L}=(1-\alpha)\cdot \mathcal{L}_{acc}+ \alpha \cdot\mathcal{L}_{flops}+\beta \cdot \mathcal{L}_{uni}
    \label{eq:loss}
\end{equation}
where $\alpha$ denotes respective loss weight for the computing efficiency, and $\beta$ controls the weight for the regularization term.

\section{Experiments}
In this section, we conduct extensive experiments to show that our model outperforms many strong baselines while significantly reducing the computation budget. We first show that our model-agnostic \ours boosts the performance of existing 2D CNN architectures (ResNet \cite{he2016deep}, EfficientNet \cite{tan2019efficientnet}) and then show our method outperforms the State-of-the-art approaches for efficient video understanding. Finally, we conduct comprehensive experiments on ablation studies and qualitative analysis to verify the effectiveness of our policy learning.

\subsection{Experimental Setup}
\label{section:setups}
\partitle{Datasets} We evaluate our approach on three large-scale action recognition datasets: ActivityNet-v1.3 ~\cite{caba2015activitynet}, FCVID(Fudan-Columbia Video Dataset)~ \cite{jiang2017exploiting} and Mini-Kinetics ~\cite{kay2017kinetics}. ActivityNet ~\cite{caba2015activitynet} is labelled with 200 action categories and contains 10,024 videos for training and 4,926 videos for validation with an average duration of 117 seconds. FCVID \cite{jiang2017exploiting} has 91,223 videos (45,611 videos for training and 45,612 videos for testing) with 239 label classes and the average length is 167 seconds. Mini-Kinetics dataset contains randomly selected 200 classes and 131,082 videos from Kinetics dataset \cite{kay2017kinetics}. We use 121,215 videos for training and 9,867 videos for testing. The average duration is 10 seconds.

\partitle{Implementation Details}
We uniformly sample $T=16$ frames from each video. During training, images are randomly cropped to $224\times 224$ patches with augmentation. At the inference stage, the images are rescaled to $256\times 256$ and center-cropped to $224\times 224$. We use four different frame resolutions ($L=4$) and three skipping strategies ($M=3$) as the action space.  
Our backbones network consists of ResNet-50 \cite{he2016deep}, ResNet-34 \cite{he2016deep}, ResNet-18 \cite{he2016deep}, and MobileNetv2 \cite{sandler2018mobilenetv2}, corresponding to the input resolutions $224\times 224$, $168\times 168$, $112\times 112$, and $84 \times 84$ respectively. The MobileNetv2 \cite{sandler2018mobilenetv2} is re-used and combined with a single-layer LSTM (with 512 hidden units) to serve as the policy network. The policy network can choose to skip 1, 2 or 4 frames. 

Policy learning in the first stage is extremely sensitive to initialization of the policy.
We observe that optimizing for both accuracy and efficiency is not effective with a randomly initialized policy. Thus, we divide The training process into 3 stages: warm-up, joint-training and fine-tuning. For warm-up, we fix the policy network and only train the backbones network (pretrained from ImageNet \cite{deng2009imagenet}) for 10 epochs with learning rate 0.02. Then the whole pipeline is jointly trained for 50 epochs with learning rate 0.001. After that, we fix the policy network parameters and fine-tune the backbones networks for 50 epochs with a lower learning rate of 0.0005. We set the initial temperature $\tau$ to 5, and gradually anneal it with an exponential decay factor of -0.045 in every epoch \cite{jang2016categorical}. We choose $\alpha=0.1$ and $\beta=0.3$ for the loss function and use SGD \cite{sutskever2013importance} with momentum 0.9 for optimization. We will make our source code and models publicly available.

\partitle{Baselines}
We compare with the following baselines and existing approaches:
\begin{itemize}[label=\textbullet]
\item UNIFORM: averages the frame-level predictions at the highest resolution $224\times 224$ from ResNet-50 as the video-level prediction.
\item LSTM: updates ResNet-50 predictions at the highest resolution $224\times 224$ by hidden states and averages all predictions as the video-level prediction.
\item RANDOM: uses our backbone framework but randomly samples policy actions from uniform distribution (instead of using learned policy distribution).
\item Multi-Scale: gathers the frame-level predictions by processing different resolutions through our backbone framework  (instead of selecting an optimal resolution with one corresponding backbone at each time step). This serves as a very strong baseline for classification, at the cost of heavy computation.
\item AdaFrame\cite{wu2019adaframe}: uses MobileNetV2/ResNet-101 as lightweight CNN/backbone.
\item LiteEval \cite{wu2019liteeval}: uses MobileNetV2/ResNet-101 as Policy Network/backbone.
\item ListenToLook(Image) \cite{gao2019listen}: we compared with a variant of their approach with only the visual modality (MobileNetv2$|$ResNet-101). We also report other results obtained by using audio data as an extra modality in Figure \ref{fig:flops_acc_curve}.  
\item SCSampler \cite{korbar2019scsampler}: as official code is not available, we re-implemented the SCSampler using AC loss as mentioned in \cite{korbar2019scsampler}. We choose MobileNetv2 as the sampler network and use ResNet-50 as the backbone. We select 10 frames out of 16 frames for prediction, as in~\cite{korbar2019scsampler}. 
\end{itemize}

\partitle{Metrics} We compute the mAP (mean average precision) and estimate the GFLOPS(gigabyte floating point operations per second) to reflect the performance for efficient video understanding. Ideally, a good system should have a high mAP with only a small amount of GFLOPS used during the inference stage. Since different baseline methods use different number of frames for classification, we calculate both GFLOPS per frame (denoted as GFLOPS/f) and GFLOPS per video (denoted as GFLOPS/V) in the following experiments.

\subsection{Main Results}
 
 \partitle{Adaptive Resolution Policy improves 2D CNN} We first compare our \ours with several simple baselines on ActivityNet and FCVID datasets to show how much performance our adaptive approach can boost in 2D convolution networks. We verify our method on both ResNet \cite{he2016deep} and EfficientNet \cite{tan2019efficientnet} to show the improvement is not limited to model architectures. As shown in Table~\ref{table:bsl}, comparing to traditional ``Uniform" and ``LSTM" methods, we save 50\% of the computation while getting a better classification performance. 
 
\begin{table}[!htbp]
\scriptsize
\def\arraystretch{1.15}
    \begin{center}
    \caption{Action recognition results (in mAP and GFLOPS) on ActivityNet-v1.3 and FCVID. Our method consistently outperforms all simple baselines}
    \label{table:bsl} 
    
  \begin{tabular}{ c| c| c | c | c | c | c | c  } 
 \hline
  \multirow{2}{*}{Approach}  & \multirow{2}{*}{Arch} &
   \multicolumn{3}{c}{ActivityNet-v1.3} & \multicolumn{3}{|c}{FCVID} \\
   \cline{3-8}
&& mAP(\%) & GFLOPS/f & GFLOPS/V& mAP(\%) & GFLOPS/f & GFLOPS/V   \\
 \hline
Uniform &\multirow{5}{*}{ResNet}  &72.5 & 4.11 & 65.76 & 81.0& 4.11 & 65.76\\
LSTM && 71.2 & 4.12 & 65.89& 81.1& 4.12 & 65.89\\
Random Policy && 65.0 & \textbf{1.04} & \textbf{16.57} & 75.3 & \textbf{1.03} & \textbf{16.49}\\
Multi-Scale && \secbest{73.5} & 6.90 & 110.43 & \textbf{81.3} & 6.90 & 110.43\\
\textbf{\ourswithoutspace}  && \textbf{73.8} & \secbest{2.09} & \secbest{33.47} &\secbest{81.3}& \secbest{2.19} & \secbest{35.12}\\
\hline
\hline
Uniform&\multirow{5}{*}{\makecell{Efficient \\ Net}}& 78.8 & 1.80 & 28.80 & 83.5& 1.80 & 28.80\\
LSTM & & 78.0 & 1.81 & 28.88 & 83.7 & 1.81 & 28.88\\
Random Policy & & 72.5 & \textbf{0.38} & \textbf{6.11} & 79.7 & \textbf{0.38} & \textbf{6.11}\\
Multi-Scale & & \secbest{79.5} & 2.35 & 37.56& \secbest{84.2} & 2.35 & 37.56\\
\textbf{\ourswithoutspace} & & \textbf{79.7} & \secbest{0.96} & \secbest{15.29} & \textbf{84.4} & \secbest{0.88} & \secbest{14.06}\\
\hline
\end{tabular}
\end{center}

\end{table}

\begin{table}[!htbp]
\scriptsize
\def\arraystretch{1.15}
  \begin{center}
  \caption{SOTA efficient methods comparison on ActivityNet-v1.3 and FCVID}
\label{table:stoa} 
  \begin{tabular}{ c | c | c | c | c | c | c } 
 \hline
  \multirow{2}{*}{Approach}  & \multicolumn{3}{c}{ActivityNet-v1.3} & \multicolumn{3}{|c}{FCVID} \\
   \cline{2-7}
& mAP(\%) & GFLOPS/f & GFLOPS/V & mAP(\%) & GFLOPS/f & GFLOPS/V  \\
 \hline
AdaFrame \cite{wu2019adaframe}      & 71.5 & 3.16 & 78.97 & 80.2 & 3.01 & 75.13\\
LiteEval \cite{wu2019liteeval}      & 72.7 & 3.80 & 95.10 & 80.0 & 3.77 & 94.30   \\
ListenToLook(Image) \cite{gao2019listen}   & 72.3 & 5.09 & 81.36 & - & - & -  \\
SCSampler \cite{korbar2019scsampler}& 72.9 & 2.62 & 41.95 & 81.0 & 2.62 & 41.95  \\
\textbf{\ourswithoutspace(ResNet)}                    & \textbf{73.8} & \textbf{2.09} & \textbf{33.47} & \textbf{81.3}& \textbf{2.19} & \textbf{35.12}\\
\hline
\hline
\textbf{\ourswithoutspace(EfficientNet)}                & \textbf{79.7} & \textbf{0.96} & \textbf{15.29} &\textbf{84.4} & \textbf{0.88} & \textbf{14.06} \\
\hline
\end{tabular}
\end{center} 
\end{table}
 
 We further show that it is the adaptively choosing resolutions and skippings that helps the most for efficient video understanding tasks. 
 Taking ResNet architecture as an example, ``Random Policy" can only reach 65.0\% mAP on ActivityNet and 75.3\% on FCVID, whereas \ours using learned policy can reach 73.8\% and 81.3\% respectively. Specifically, ``Multi-Scale" can be a very strong baseline because it gathers all the predictions from multi-scale inputs through multiple backbones. It is noticeable that \ourswithoutspace's classification performance is comparable to the ``Multi-Scale" baseline, while using 70\% less computation. One possible explanation is that there exist noisy and misleading frames in the videos, and \ours learns to skip those frames and uses the rest of the frames for prediction. Similar conclusion can also be drawn from using EfficientNet architectures, which shows our approach is model-agnostic. 

\partitle{Adaptive Resolution Policy outperforms state-of-the-art methods} We compare the performance of \ours with several state-of-the-art methods on ActivityNet and FCVID in Table \ref{table:stoa}. The result section of the table is divided into two parts. The upper part contains all the methods using Residual Network architecture, whereas the lower part shows the best result we have achieved by using the latest EfficientNet \cite{tan2019efficientnet} architecture. Usually it is hard to improve the classification accuracy while maintaining a low computation cost, but our ``\ourswithoutspace(ResNet)" outperforms all the state-of-the-art methods in terms of mAP scores, frame-level GFLOPS and video-level GFLOPS. Our method achieves 73.8\% mAP on ActivityNet and 81.3\% mAP on FCVID while using 17\% $\sim$ 64\% less computation budgets compared with other approaches. This shows the power of our adaptive resolution learning approach in efficient video understanding tasks. When integrated with EfficientNet \cite{tan2019efficientnet}, our ``\ourswithoutspace(EfficientNet)" further gains 5.9\% in mAP on ActivityNet and 3.1\% on FCVID, with 54\%$\sim$60\% less computation compared to ``\ourswithoutspace(ResNet)". Since there is no published result using EfficientNet for efficient video understanding, these results can serve as the new baselines for future research.

Figure~\ref{fig:flops_acc_curve} illustrates the GFLOPS-mAP curve on ActivityNet dataset, where our \ours obtains significant computational efficiency and action recognition accuracy with much fewer GFLOPS than other baseline methods.  We quote the reported results on MultiAgent \cite{wu2019multi}, AdaFrame\cite{wu2019adaframe} and ListenToLook\cite{gao2019listen}  (here ``(IA$|$R)" and ``(MN$|$R)" are short for ``(Image-Audio$|$ResNet-101)" and ``(MobileNetV2$|$ResNet-101)" mentioned in \cite{gao2019listen}). The results of LiteEval \cite{wu2019liteeval} are generated through the codes shared by the authors, and the results of SCSampler \cite{korbar2019scsampler} are obtained by our re-implementation following their reported details. ListenToLook (IA$|$R) denotes models using both visual and audio data as inputs. 
Given the same ResNet architectural family, our approach achieves substantial improvement compared to the best competitors, demonstrating the superiority of our method. Additionally, our best performing model, which employs EfficientNet \cite{tan2019efficientnet} architecture, yields more than $5\%$ improvement in mAP at the same computation budgets. It shows that our approach of adaptively selecting proper resolutions on a per frame basis is able to yield significant savings in computation budget and to improve recognition precision.
\begin{SCfigure}[][t]
\centering
\pgfplotsset{
every axis plot/.append style={line width=1pt},
tick label style={font=\small},
label style={font=\small},
legend style={font=\scriptsize, at={(1,0)},anchor=south east}
}

\begin{tikzpicture}
    \begin{axis}[
      height=8.8cm,
	  width=8.8cm,
      xlabel=GFLOPS/Video,
      ylabel=mean Average Precision (\%),
      xmin=0,
      xmax=310,
      ymin=59,
      ymax=82,
      legend style={nodes={inner sep=1.5pt,text depth=0.0em}}]
    ]
    \input{my_tikz/data/data_multiagent}
    \input{my_tikz/data/data_liteeval}
    \input{my_tikz/data/data_adaframe10}
    \input{my_tikz/data/data_adaframe5}
    \input{my_tikz/data/data_ltl_mobilenet_resnet}
    \input{my_tikz/data/data_ltl_image_audio_resnet}
    \input{my_tikz/data/data_ltl_image_audio_image_audio}
    \input{my_tikz/data/data_scsampler}
    \input{my_tikz/data/data_ours_resnet}
    \input{my_tikz/data/data_ours_effnet}
    \legend{
    MultiAgent \cite{wu2019multi}\\
    LiteEval \cite{wu2019liteeval}\\
    AdaFrame10 \cite{wu2019adaframe}\\
    AdaFrame5 \cite{wu2019adaframe}\\
    ListenToLook(MN$|$R) \cite{gao2019listen}\\
    ListenToLook(IA$|$R) \cite{gao2019listen}\\
    ListenToLook(IA$|$IA) \cite{gao2019listen}\\
    SCSampler \cite{korbar2019scsampler}\\
    Ours (ResNet) \\
    Ours (EfficientNet)\\
    }
    \end{axis}
\end{tikzpicture}
\caption{Comparisons with state-of-the-art alternatives on ActivityNet dataset. Our proposed \ours obtains the best recognition accuracy with much fewer GFLOPS than the compared methods. We directly quote the numbers reported in published papers when possible and compare the mAP against the average GFLOPs per test video. See text for more details.}
   \label{fig:flops_acc_curve}
\end{SCfigure}
\begin{table}[tbp]
\begin{center}
\end{center}
\scriptsize
\def\arraystretch{1.15}
  \begin{center}
  \caption{Results for video classification on Mini-Kinetics dataset} 
  \label{table:minik-exp}
  \begin{tabular}{ c | c | c | c  } 
 \hline
  \multirow{2}{*}{Approach}  & \multicolumn{3}{c}{Mini-Kinetics}\\
   \cline{2-4}
& Top1(\%) & GFLOPS/f & GFLOPS/V  \\
 \hline
LiteEval \cite{wu2019liteeval}      & 61.0 & 3.96 & 99.00  \\
SCSampler \cite{korbar2019scsampler}& 70.8 & 2.62 & 41.95     \\
\textbf{\ourswithoutspace(ResNet)}  & \textbf{71.7}& \textbf{2.00} & \textbf{32.00} \\
\hline
\hline
\textbf{\ourswithoutspace(EfficientNet)} & \textbf{74.8}& \textbf{1.02} & \textbf{16.32}  \\
\hline
\end{tabular}
\end{center}
\end{table}
\partitle{Further Experiment on Mini-Kinetics} 
To further test the capability of our method, we conduct experiments on Mini-Kinetics dataset. Compared with the recent methods LiteEval \cite{wu2019liteeval} and SCSampler \cite{korbar2019scsampler}, our method achieves better Top-1 accuracy and the computation cost is reduced with noticeable margin. In brief, our method consistently outperform the existing methods in terms of accuracy and speed on different datasets, which implies our \ours provides an effective framework for various action recognition applications.

\begin{table}[tbp]
\scriptsize
\def\arraystretch{1.05}
\begin{center}
\caption{Results of different policy settings on ActivityNet-v1.3}
\label{table:policy}
  \begin{tabular}{c | c | c | c} 
 \hline
  Policy Settings & mAP(\%) & GFLOPS/f & GFLOPS/V\\
 \hline
Uniform & 72.5 & 4.11 & 65.76\\
LSTM & 71.2 & 4.12 & 65.89\\
Resolution Only & 73.4 & 2.13 & 34.08\\
Skipping Only& 72.7 & 2.21 & 34.90\\
\textbf{Resolution+Skipping} &  \textbf{73.8} & \textbf{2.09} & \textbf{33.47}\\
\hline
\end{tabular}
\end{center}
  \begin{center}
\caption{Results of different losses on ActivityNet-v1.3}
\label{table:loss}
  \begin{tabular}{ c | c |  c | c | c | c} 
 \hline
   Losses & $\alpha$ & $\beta$ & mAP(\%) & GFLOPS/f & GFLOPS/V \\
 \hline
 Acc &$\,\,$0.0$\,\,$&$\,\,$0.0$\,\,$&\textbf{74.5} & 3.75 & 60.06\\
 Acc+Eff & 0.1 & 0.0 &73.8 & 2.28 & 36.48\\
 \textbf{Acc+Eff+Uni}& 0.1 & 0.3 & 73.8 & \textbf{2.09} & \textbf{33.47}\\
\hline
\end{tabular}
 \end{center}
  \begin{center}
\caption{Results of different training strategies on ActivityNet-v1.3}
\label{table:training}
  \begin{tabular}{c c c | c | c | c} 
 \hline
  \multicolumn{3}{c|}{Training Strategy} & \multirow{2}{*}{mAP(\%)} & \multirow{2}{*}{GFLOPS/f} & \multirow{2}{*}{GFLOPS/V} \\
 \cline{1-3}
 Warm-Up& $\,\,\,\,$ Joint $\,\,\,\,$ &Finetuning & & & \\
 \hline
 \xmark&\cmark&\xmark & 67.1 & 1.16 & 17.86\\
\cmark&\cmark&\xmark & 73.3 & 2.03 & 32.40\\
\cmark&\cmark&\cmark &  73.8 & 2.09 & 33.47\\
\hline
\end{tabular}
\end{center}
\end{table}

\subsection{Ablation Studies}
\partitle{Effectiveness of choosing resolution and skipping} Here we inspect how each type of operation enhances the efficient video understanding. We define three different action spaces: ``Resolution Only" (the policy network can only choose different resolutions), ``Skipping Only"(the policy network can only decide how many frames to skip) and ``Resolution+Skipping". We follow the same training procedures as illustrated in Section \ref{section:setups} and evaluate each approach on ActivityNet dataset. We adjust the training loss to keep their GFLOPS at the same level and we only compare the differences in classification performances. As shown in Table \ref{table:policy}, comparing with baseline methods (``Uniform" and ``LSTM"), they all improve the performance, and the best strategy is to combine skippings and choosing resolutions. 
Intuitively, skipping can be seen as ``choosing zero resolution" for the current frame, hence gives more flexibility in decision-making.

\partitle{Trade-off between accuracy and efficiency} As discussed in Section~\ref{section:loss}, hyper-parameters $\alpha$ and $\beta$ in Equation~\ref{eq:loss} affect the classification performance, efficiency and policy distribution. Here we train our model using 3 different weighted combinations: ``Acc" (only using accuracy-related loss), ``Acc+Eff"(using accuracy and efficiency losses) and ``Acc+Eff+Uni"(using all the losses).  As shown in Table~\ref{table:loss},  training with ``Acc" will achieve the highest mAP, but the computation cost will be similar to ``Uniform" method (GFLOPS/V=65.76). Adding the efficiency loss term will decrease the computation cost drastically, whereas training with ``Acc+Eff+Uni" will drop the GFLOPS even further. One reason is that the network tends to skip more frames in the inference stage. Finally, we use hyper-parameters $\alpha=0.1,\,\beta=0.3$ in our training.

\partitle{Different training strategies} We explore several strategies for training the adaptive learning framework. As shown in Table~\ref{table:training}, the best practice comes from ``Warm-Up+Joint+Finetuning" so we adopt it in training our models.

\begin{figure}[!tbp]
    \centering
    \includegraphics[width=.865\linewidth]{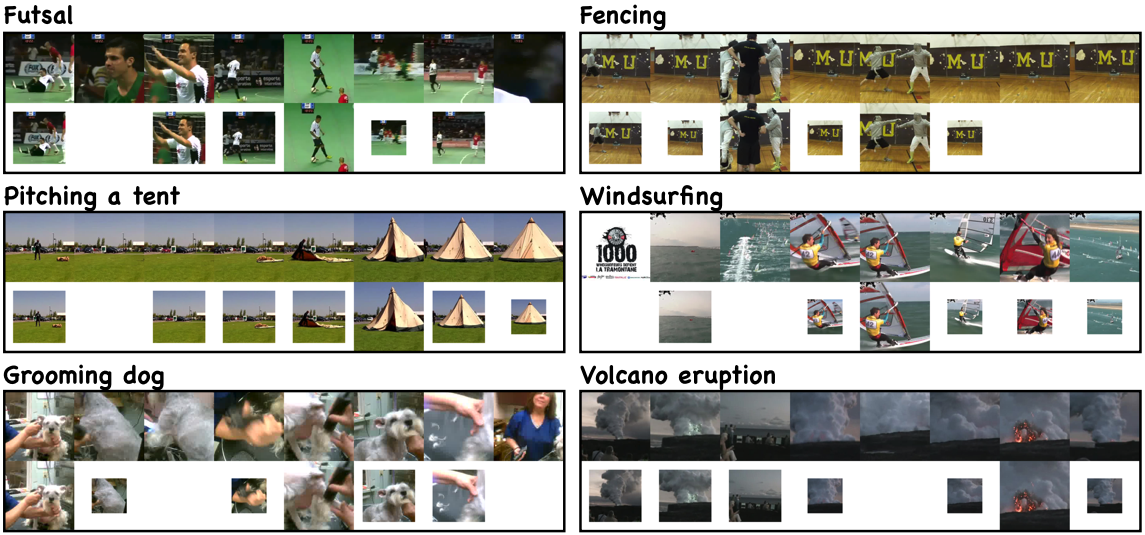}
    \caption{Qualitative examples from ActivityNet and FCVID. We uniformly sample 8 frames per video and \ours chooses the proper resolutions or skipping. Relevant frames are kept in original resolution whereas non-informative frames are resized to lower resolution or skipped for computation efficiency.}
    \label{fig:screenshot}
\end{figure}

\subsection{Qualitative Analysis}

An intuitive view of how \ours achieves efficiency is shown in Figure~\ref{fig:screenshot}. We conduct experiments on ActivityNet-v1.3 and FCVID testing sets. Videos are uniformly sampled in 8 frames. The upper row of each example shows original input frames, and the lower row shows the frames processed by our policy network for predictions. \ours keeps the most indicative frames (e.g. Futsal and Fencing) in original resolution and resizes or skips frames that are irrelevant or in low quality (blurriness). After being confident about the predictions, \ours will avoid using the original resolution even if informative contents appear again (e.g. Pitching a tent/Windsurfing). The last two examples show that \ours is able to capture both object-interaction (clipper-dog) and background changes.

\section{Conclusion}
We have demonstrated the power of adaptive resolution learning on a per frame basis for efficient video action recognition. Comprehensive experiments show that our method can work in a full range of accuracy-speed operating points, from a version that is both faster and more accurate than comparable visual-only models to a new, state-of-the-art accuracy-throughput version based on the EfficientNet \cite{tan2019efficientnet} architecture.
The proposed learning framework is model-agnostic, which allows applications to various sophisticated backbone networks and the idea can be generally adopted to explore other complex video understanding tasks.

\section*{Acknowledgement}

This work is supported by IARPA via DOI/IBC contract number D17PC00341. The U.S. Government is authorized to reproduce and distribute reprints for Governmental purposes notwithstanding any copyright annotation thereon. This work is partly supported by the MIT-IBM Watson AI Lab.

\nottoggle{Arxiv}{}{
\parclaim{Disclaimer} The views and conclusions contained herein are those of the authors and should not be interpreted as necessarily representing the official policies or endorsements, either expressed or implied, of IARPA, DOI/IBC, or the U.S. Government.
}
\bibliography{z7_refere}

\begin{thebibliography}{10}
\providecommand{\url}[1]{\texttt{#1}}
\providecommand{\urlprefix}{URL }
\providecommand{\doi}[1]{https://doi.org/#1}

\bibitem{adelson1984pyramid}
Adelson, E.H., Anderson, C.H., Bergen, J.R., Burt, P.J., Ogden, J.M.: Pyramid
  methods in image processing. RCA engineer  \textbf{29}(6),  33--41 (1984)

\bibitem{araujo2018training}
Araujo, A., Negrevergne, B., Chevaleyre, Y., Atif, J.: Training compact deep
  learning models for video classification using circulant matrices. In:
  Proceedings of the European Conference on Computer Vision (ECCV). pp.~0--0
  (2018)

\bibitem{bengio2015conditional}
Bengio, E., Bacon, P.L., Pineau, J., Precup, D.: Conditional computation in
  neural networks for faster models. arXiv preprint arXiv:1511.06297  (2015)

\bibitem{bengio2013estimating}
Bengio, Y., L{\'e}onard, N., Courville, A.: Estimating or propagating gradients
  through stochastic neurons for conditional computation. arXiv preprint
  arXiv:1308.3432  (2013)

\bibitem{caba2015activitynet}
Caba~Heilbron, F., Escorcia, V., Ghanem, B., Carlos~Niebles, J.: Activitynet: A
  large-scale video benchmark for human activity understanding. In: Proceedings
  of the ieee conference on computer vision and pattern recognition. pp.
  961--970 (2015)

\bibitem{cai2016unified}
Cai, Z., Fan, Q., Feris, R.S., Vasconcelos, N.: A unified multi-scale deep
  convolutional neural network for fast object detection. In: European
  conference on computer vision. pp. 354--370. Springer (2016)

\bibitem{carreira2017quo}
Carreira, J., Zisserman, A.: Quo vadis, action recognition? a new model and the
  kinetics dataset. In: proceedings of the IEEE Conference on Computer Vision
  and Pattern Recognition. pp. 6299--6308 (2017)

\bibitem{chen2018big}
Chen, C.F., Fan, Q., Mallinar, N., Sercu, T., Feris, R.: Big-little net: An
  efficient multi-scale feature representation for visual and speech
  recognition. arXiv preprint arXiv:1807.03848  (2018)

\bibitem{chen2015compressing}
Chen, W., Wilson, J., Tyree, S., Weinberger, K., Chen, Y.: Compressing neural
  networks with the hashing trick. In: International conference on machine
  learning. pp. 2285--2294 (2015)

\bibitem{cheron2015p}
Ch{\'e}ron, G., Laptev, I., Schmid, C.: P-cnn: Pose-based cnn features for
  action recognition. In: Proceedings of the IEEE international conference on
  computer vision. pp. 3218--3226 (2015)

\bibitem{deng2009imagenet}
Deng, J., Dong, W., Socher, R., Li, L.J., Li, K., Fei-Fei, L.: Imagenet: A
  large-scale hierarchical image database. In: 2009 IEEE conference on computer
  vision and pattern recognition. pp. 248--255. Ieee (2009)

\bibitem{donahue2015long}
Donahue, J., Anne~Hendricks, L., Guadarrama, S., Rohrbach, M., Venugopalan, S.,
  Saenko, K., Darrell, T.: Long-term recurrent convolutional networks for
  visual recognition and description. In: Proceedings of the IEEE conference on
  computer vision and pattern recognition. pp. 2625--2634 (2015)

\bibitem{dong2017more}
Dong, X., Huang, J., Yang, Y., Yan, S.: More is less: A more complicated
  network with less inference complexity. In: Proceedings of the IEEE
  Conference on Computer Vision and Pattern Recognition. pp. 5840--5848 (2017)

\bibitem{fan2018watching}
Fan, H., Xu, Z., Zhu, L., Yan, C., Ge, J., Yang, Y.: Watching a small portion
  could be as good as watching all: Towards efficient video classification. In:
  IJCAI International Joint Conference on Artificial Intelligence (2018)

\bibitem{fan2019more}
Fan, Q., Chen, C.F.R., Kuehne, H., Pistoia, M., Cox, D.: More is less: Learning
  efficient video representations by big-little network and depthwise temporal
  aggregation. In: Advances in Neural Information Processing Systems. pp.
  2261--2270 (2019)

\bibitem{feichtenhofer2019slowfast}
Feichtenhofer, C., Fan, H., Malik, J., He, K.: Slowfast networks for video
  recognition. In: Proceedings of the IEEE International Conference on Computer
  Vision. pp. 6202--6211 (2019)

\bibitem{feichtenhofer2017spatiotemporal}
Feichtenhofer, C., Pinz, A., Wildes, R.P.: Spatiotemporal multiplier networks
  for video action recognition. In: Proceedings of the IEEE conference on
  computer vision and pattern recognition. pp. 4768--4777 (2017)

\bibitem{figurnov2017spatially}
Figurnov, M., Collins, M.D., Zhu, Y., Zhang, L., Huang, J., Vetrov, D.,
  Salakhutdinov, R.: Spatially adaptive computation time for residual networks.
  In: Proceedings of the IEEE Conference on Computer Vision and Pattern
  Recognition. pp. 1039--1048 (2017)

\bibitem{gao2018dynamic}
Gao, M., Yu, R., Li, A., Morariu, V.I., Davis, L.S.: Dynamic zoom-in network
  for fast object detection in large images. In: Proceedings of the IEEE
  Conference on Computer Vision and Pattern Recognition. pp. 6926--6935 (2018)

\bibitem{gao2019listen}
Gao, R., Oh, T.H., Grauman, K., Torresani, L.: Listen to look: Action
  recognition by previewing audio. arXiv preprint arXiv:1912.04487  (2019)

\bibitem{gkioxari2015finding}
Gkioxari, G., Malik, J.: Finding action tubes. In: Proceedings of the IEEE
  conference on computer vision and pattern recognition. pp. 759--768 (2015)

\bibitem{glynn1990likelihood}
Glynn, P.W.: Likelihood ratio gradient estimation for stochastic systems.
  Communications of the ACM  \textbf{33}(10),  75--84 (1990)

\bibitem{graves2016adaptive}
Graves, A.: Adaptive computation time for recurrent neural networks. arXiv
  preprint arXiv:1603.08983  (2016)

\bibitem{guo2019spottune}
Guo, Y., Shi, H., Kumar, A., Grauman, K., Rosing, T., Feris, R.: Spottune:
  transfer learning through adaptive fine-tuning. In: Proceedings of the IEEE
  Conference on Computer Vision and Pattern Recognition. pp. 4805--4814 (2019)

\bibitem{hara2018can}
Hara, K., Kataoka, H., Satoh, Y.: Can spatiotemporal 3d cnns retrace the
  history of 2d cnns and imagenet? In: Proceedings of the IEEE conference on
  Computer Vision and Pattern Recognition. pp. 6546--6555 (2018)

\bibitem{he2016deep}
He, K., Zhang, X., Ren, S., Sun, J.: Deep residual learning for image
  recognition. In: Proceedings of the IEEE conference on computer vision and
  pattern recognition. pp. 770--778 (2016)

\bibitem{howard2017mobilenets}
Howard, A.G., Zhu, M., Chen, B., Kalenichenko, D., Wang, W., Weyand, T.,
  Andreetto, M., Adam, H.: Mobilenets: Efficient convolutional neural networks
  for mobile vision applications. arXiv preprint arXiv:1704.04861  (2017)

\bibitem{iandola2016squeezenet}
Iandola, F.N., Han, S., Moskewicz, M.W., Ashraf, K., Dally, W.J., Keutzer, K.:
  Squeezenet: Alexnet-level accuracy with 50x fewer parameters and< 0.5 mb
  model size. arXiv preprint arXiv:1602.07360  (2016)

\bibitem{jang2016categorical}
Jang, E., Gu, S., Poole, B.: Categorical reparameterization with
  gumbel-softmax. arXiv preprint arXiv:1611.01144  (2016)

\bibitem{jiang2017exploiting}
Jiang, Y.G., Wu, Z., Wang, J., Xue, X., Chang, S.F.: Exploiting feature and
  class relationships in video categorization with regularized deep neural
  networks. IEEE transactions on pattern analysis and machine intelligence
  \textbf{40}(2),  352--364 (2017)

\bibitem{karpathy2014large}
Karpathy, A., Toderici, G., Shetty, S., Leung, T., Sukthankar, R., Fei-Fei, L.:
  Large-scale video classification with convolutional neural networks. In:
  Proceedings of the IEEE conference on Computer Vision and Pattern
  Recognition. pp. 1725--1732 (2014)

\bibitem{kay2017kinetics}
Kay, W., Carreira, J., Simonyan, K., Zhang, B., Hillier, C., Vijayanarasimhan,
  S., Viola, F., Green, T., Back, T., Natsev, P., et~al.: The kinetics human
  action video dataset. arXiv preprint arXiv:1705.06950  (2017)

\bibitem{korbar2019scsampler}
Korbar, B., Tran, D., Torresani, L.: Scsampler: Sampling salient clips from
  video for efficient action recognition. In: Proceedings of the IEEE
  International Conference on Computer Vision. pp. 6232--6242 (2019)

\bibitem{laptev2008learning}
Laptev, I., Marszalek, M., Schmid, C., Rozenfeld, B.: Learning realistic human
  actions from movies. In: 2008 IEEE Conference on Computer Vision and Pattern
  Recognition. pp.~1--8. IEEE (2008)

\bibitem{li2016pruning}
Li, H., Kadav, A., Durdanovic, I., Samet, H., Graf, H.P.: Pruning filters for
  efficient convnets. arXiv preprint arXiv:1608.08710  (2016)

\bibitem{lin2019tsm}
Lin, J., Gan, C., Han, S.: Tsm: Temporal shift module for efficient video
  understanding. In: Proceedings of the IEEE International Conference on
  Computer Vision. pp. 7083--7093 (2019)

\bibitem{lin2017feature}
Lin, T.Y., Doll{\'a}r, P., Girshick, R., He, K., Hariharan, B., Belongie, S.:
  Feature pyramid networks for object detection. In: Proceedings of the IEEE
  conference on computer vision and pattern recognition. pp. 2117--2125 (2017)

\bibitem{mcgill2017deciding}
McGill, M., Perona, P.: Deciding how to decide: Dynamic routing in artificial
  neural networks. In: Proceedings of the 34th International Conference on
  Machine Learning-Volume 70. pp. 2363--2372 (2017)

\bibitem{monfort2019moments}
Monfort, M., Andonian, A., Zhou, B., Ramakrishnan, K., Bargal, S.A., Yan, T.,
  Brown, L., Fan, Q., Gutfreund, D., Vondrick, C., et~al.: Moments in time
  dataset: one million videos for event understanding. IEEE transactions on
  pattern analysis and machine intelligence  \textbf{42}(2),  502--508 (2019)

\bibitem{monfort2019multi}
Monfort, M., Ramakrishnan, K., Andonian, A., McNamara, B.A., Lascelles, A.,
  Pan, B., Gutfreund, D., Feris, R., Oliva, A.: Multi-moments in time: Learning
  and interpreting models for multi-action video understanding. arXiv preprint
  arXiv:1911.00232  (2019)

\bibitem{najibi2019autofocus}
Najibi, M., Singh, B., Davis, L.S.: Autofocus: Efficient multi-scale inference.
  In: Proceedings of the IEEE International Conference on Computer Vision. pp.
  9745--9755 (2019)

\bibitem{pedersoli2015coarse}
Pedersoli, M., Vedaldi, A., Gonzalez, J., Roca, X.: A coarse-to-fine approach
  for fast deformable object detection. Pattern Recognition  \textbf{48}(5),
  1844--1853 (2015)

\bibitem{perona1990scale}
Perona, P., Malik, J.: Scale-space and edge detection using anisotropic
  diffusion. IEEE Transactions on pattern analysis and machine intelligence
  \textbf{12}(7),  629--639 (1990)

\bibitem{piergiovanni2019tiny}
Piergiovanni, A., Angelova, A., Ryoo, M.S.: Tiny video networks. arXiv preprint
  arXiv:1910.06961  (2019)

\bibitem{sandler2018mobilenetv2}
Sandler, M., Howard, A., Zhu, M., Zhmoginov, A., Chen, L.C.: Mobilenetv2:
  Inverted residuals and linear bottlenecks. In: Proceedings of the IEEE
  conference on computer vision and pattern recognition. pp. 4510--4520 (2018)

\bibitem{simonyan2014two}
Simonyan, K., Zisserman, A.: Two-stream convolutional networks for action
  recognition in videos. In: Advances in neural information processing systems.
  pp. 568--576 (2014)

\bibitem{sutskever2013importance}
Sutskever, I., Martens, J., Dahl, G., Hinton, G.: On the importance of
  initialization and momentum in deep learning. In: International conference on
  machine learning. pp. 1139--1147 (2013)

\bibitem{tan2019efficientnet}
Tan, M., Le, Q.V.: Efficientnet: Rethinking model scaling for convolutional
  neural networks. arXiv preprint arXiv:1905.11946  (2019)

\bibitem{tran2015learning}
Tran, D., Bourdev, L., Fergus, R., Torresani, L., Paluri, M.: Learning
  spatiotemporal features with 3d convolutional networks. In: Proceedings of
  the IEEE international conference on computer vision. pp. 4489--4497 (2015)

\bibitem{tran2019video}
Tran, D., Wang, H., Torresani, L., Feiszli, M.: Video classification with
  channel-separated convolutional networks. In: Proceedings of the IEEE
  International Conference on Computer Vision. pp. 5552--5561 (2019)

\bibitem{tran2018closer}
Tran, D., Wang, H., Torresani, L., Ray, J., LeCun, Y., Paluri, M.: A closer
  look at spatiotemporal convolutions for action recognition. In: Proceedings
  of the IEEE conference on Computer Vision and Pattern Recognition. pp.
  6450--6459 (2018)

\bibitem{veit2018convolutional}
Veit, A., Belongie, S.: Convolutional networks with adaptive inference graphs.
  In: Proceedings of the European Conference on Computer Vision (ECCV). pp.
  3--18 (2018)

\bibitem{wang2011action}
Wang, H., Kl{\"a}ser, A., Schmid, C., Liu, C.L.: Action recognition by dense
  trajectories. In: CVPR 2011. pp. 3169--3176. IEEE (2011)

\bibitem{wang2016temporal}
Wang, L., Xiong, Y., Wang, Z., Qiao, Y., Lin, D., Tang, X., Van~Gool, L.:
  Temporal segment networks: Towards good practices for deep action
  recognition. In: European conference on computer vision. pp. 20--36. Springer
  (2016)

\bibitem{wang2018non}
Wang, X., Girshick, R., Gupta, A., He, K.: Non-local neural networks. In:
  Proceedings of the IEEE conference on computer vision and pattern
  recognition. pp. 7794--7803 (2018)

\bibitem{wang2018skipnet}
Wang, X., Yu, F., Dou, Z.Y., Darrell, T., Gonzalez, J.E.: Skipnet: Learning
  dynamic routing in convolutional networks. In: Proceedings of the European
  Conference on Computer Vision (ECCV). pp. 409--424 (2018)

\bibitem{wen2017coordinating}
Wen, W., Xu, C., Wu, C., Wang, Y., Chen, Y., Li, H.: Coordinating filters for
  faster deep neural networks. In: Proceedings of the IEEE International
  Conference on Computer Vision. pp. 658--666 (2017)

\bibitem{williams1992simple}
Williams, R.J.: Simple statistical gradient-following algorithms for
  connectionist reinforcement learning. Machine learning  \textbf{8}(3-4),
  229--256 (1992)

\bibitem{wu2019multi}
Wu, W., He, D., Tan, X., Chen, S., Wen, S.: Multi-agent reinforcement learning
  based frame sampling for effective untrimmed video recognition. In:
  Proceedings of the IEEE International Conference on Computer Vision. pp.
  6222--6231 (2019)

\bibitem{wu2018blockdrop}
Wu, Z., Nagarajan, T., Kumar, A., Rennie, S., Davis, L.S., Grauman, K., Feris,
  R.: Blockdrop: Dynamic inference paths in residual networks. In: Proceedings
  of the IEEE Conference on Computer Vision and Pattern Recognition. pp.
  8817--8826 (2018)

\bibitem{wu2019liteeval}
Wu, Z., Xiong, C., Jiang, Y.G., Davis, L.S.: Liteeval: A coarse-to-fine
  framework for resource efficient video recognition. In: Advances in Neural
  Information Processing Systems. pp. 7778--7787 (2019)

\bibitem{wu2019adaframe}
Wu, Z., Xiong, C., Ma, C.Y., Socher, R., Davis, L.S.: Adaframe: Adaptive frame
  selection for fast video recognition. In: Proceedings of the IEEE Conference
  on Computer Vision and Pattern Recognition. pp. 1278--1287 (2019)

\bibitem{yeung2016end}
Yeung, S., Russakovsky, O., Mori, G., Fei-Fei, L.: End-to-end learning of
  action detection from frame glimpses in videos. In: Proceedings of the IEEE
  Conference on Computer Vision and Pattern Recognition. pp. 2678--2687 (2016)

\bibitem{zhang2018shufflenet}
Zhang, X., Zhou, X., Lin, M., Sun, J.: Shufflenet: An extremely efficient
  convolutional neural network for mobile devices. In: Proceedings of the IEEE
  conference on computer vision and pattern recognition. pp. 6848--6856 (2018)

\bibitem{zhou2018temporal}
Zhou, B., Andonian, A., Oliva, A., Torralba, A.: Temporal relational reasoning
  in videos. In: Proceedings of the European Conference on Computer Vision
  (ECCV). pp. 803--818 (2018)

\end{thebibliography}
\renewcommand\thesection{\Alph{section}}
\setcounter{figure}{0}
\setcounter{table}{0}

\title{Adaptive Resolution for Efficient Action Recognition (Supplementary)}
\author{}
\institute{}
\maketitle

\begin{table}
    \begin{center}
    \caption{\small \textbf{Supplementary Material Overview}}
        \label{table:Appendix_Content}
        \resizebox{0.6\linewidth}{!}{
        \begin{tabular}{c|c}
             \Xhline{3\arrayrulewidth} 
             Section & Content \\
             \hline
             A & Details on Mini-Kinetics Dataset \\
             B & GFLOPS Estimation \\
             C & Policy Distributions \\
             D & Qualitative Analysis \\
             E & RL vs Gumbel Softmax in Policy Learning \\
            \Xhline{3\arrayrulewidth} 
        \end{tabular}}
    \end{center}
 
\end{table}

\section{Mini-Kinetics} Kinetics is a large dataset containing 400 action classes and 240K training videos that are collected from YouTube. Since the full Kinetics dataset is quite large, we have created a smaller dataset that we call Mini-Kinetics by randomly selecting half of the categories of Kinetics-400 dataset. The mini-Kinetics dataset contains 121K videos for training and 10K videos for testing, with each video lasting 6-10 seconds. We will make the splits publicly available to enable future comparisons.

\section{GFLOPS Estimation}
\begin{table}[!htbp]
\def\arraystretch{1.3}
  \caption{GFLOPS for different backbones and resolutions}
  \begin{center}
  \begin{tabular}{ c | c | c | c } 
 \hline
  
Network & Resolution & GFLOPS & Feature Dim\\
\hline
MobileNet$-$v2 & 84$\times$84 & 0.0529 & 1280\\
ResNet$-$18 & 112$\times$112 & 0.4683 & 512\\
ResNet$-$34 & 168$\times$168 & 2.2490 & 512\\
ResNet$-$50 & 224$\times$224 & 4.1103 & 2048\\
EfficientNet$-$b0 & 112$\times$112 & 0.0975 & 1280\\ 
EfficientNet$-$b1 & 168$\times$168 & 0.3937 & 1280\\ 
EfficientNet$-$b3 & 224$\times$224 & 1.8000 & 1536\\ 

\hline
\end{tabular}
\end{center} 
\label{table:appendix-1} 
\end{table}

To estimate the overall GFLOPS for our framework, we compute a weighted sum based on online policy distribution and an offline GFLOPS look up table. The method to compute online policy distribution is summarized in Equation 11. To generate the look up table for GFLOPS with respect to different modules and resolutions, we first  instantiate the specific network and then use THOP (\url{https://pypi.org/project/thop/}) to measure the GFLOPS. The example code snippet for computing the FLOPS for ResNet$-$50 at $224\times 224$ frame resolution is given below. 

\begin{lstlisting}[language=Python]
import torch, torchvision, thop
model = getattr(torchvision.models, "resnet50")(True)
data =(torch.randn(1, 3, 224, 224),)
flops, _ = thop.profile(model, inputs=data)
\end{lstlisting}

Table \ref{table:appendix-1} presents all the results we need for computing GFLOPS. The GFLOPS for LSTM is approximated by ``square of the input feature dimension", since we only need matrix-vector multiplications. Note that when feature dimension is around $1000\sim2000$, this value is normally smaller than 0.01, and hence negligible to other operations.

\section{Distributions}
\begin{figure}
\subfloat[Overall policy distribution on ActivityNet, FCVID and Mini-Kinetics]{\includegraphics[width = .999\linewidth]{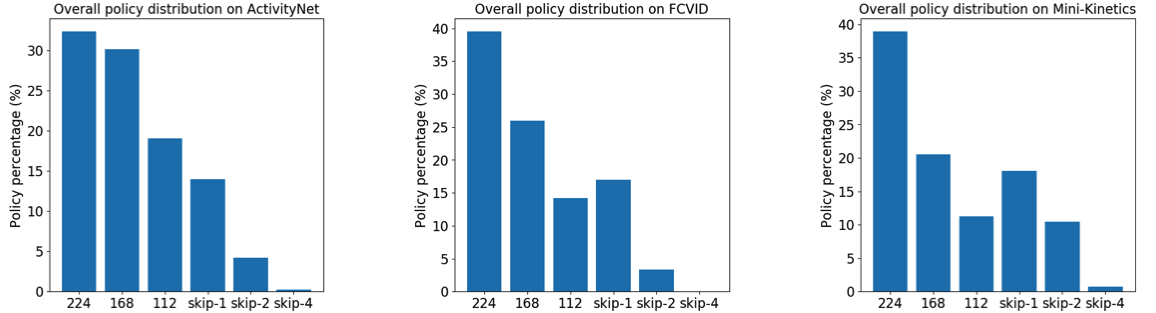}} \\
\vspace{2em}
\subfloat[Relative high resolution usage on ActivityNet, FCVID and Mini-Kinetics]{\includegraphics[width = .999\linewidth]{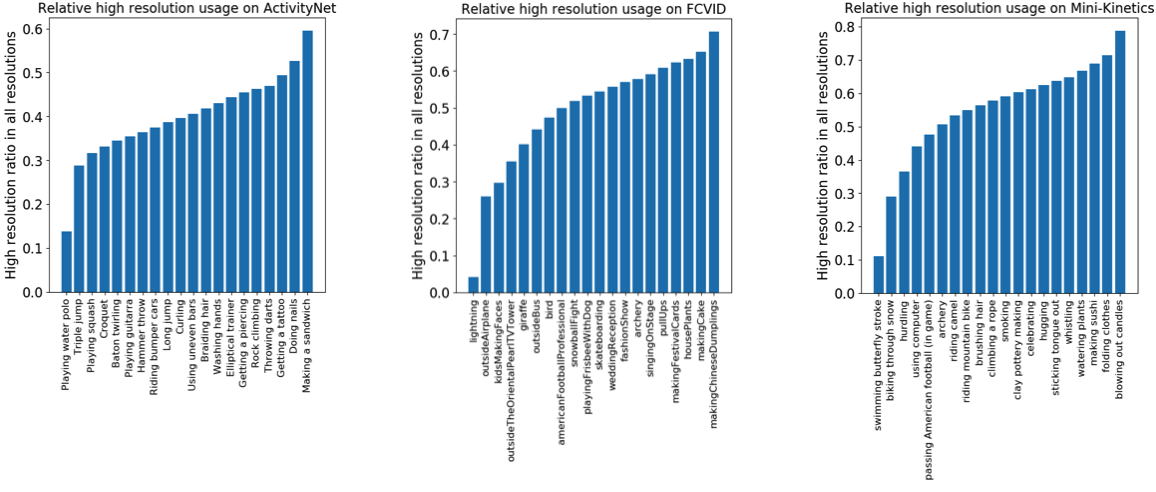}} \\
\vspace{2em}
\subfloat[Resolution usage on ActivityNet, FCVID and Mini-Kinetics]{\includegraphics[width = .999\linewidth]{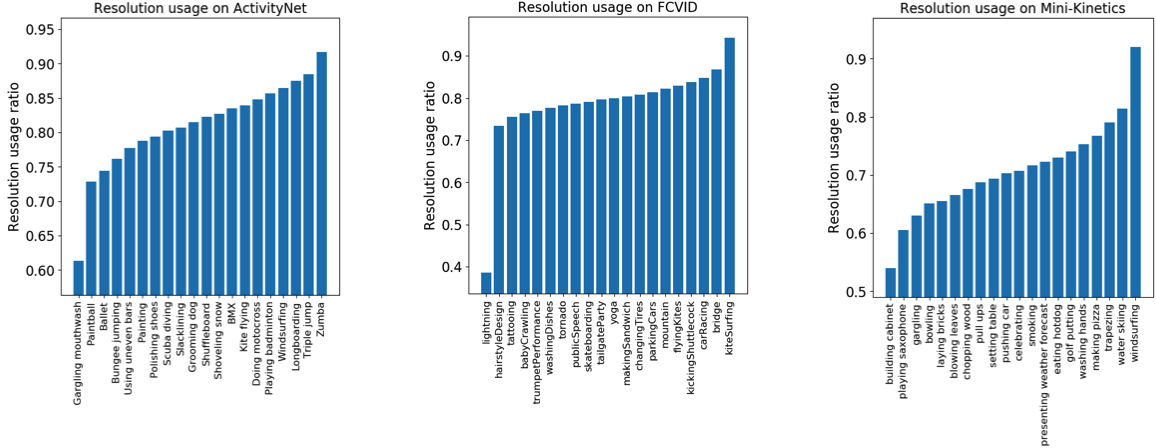}} \\
\vspace{-1.5em}
\caption{Dataset-specific and category-specific policy usage for AR-Net(ResNet).}
\label{fig:supple-1}
\end{figure}

Figure \ref{fig:supple-1} shows the dataset-specific and category-specific policy usages for ``AR-Net(ResNet)". Videos are uniformly sampled in 8 frames. We present policy distribution (choosing 224$\times$224/168$\times$168/112$\times$112 resolution or skipping 1/2/4 frames) in Figure \ref{fig:supple-1}(a), present a subset of classes sorted in relative high resolution usage (ratio of ``choosing 224$\times$224" over ``choosing 224$\times$224/168$\times$168/ 112$\times$112") in Figure \ref{fig:supple-1}(b) and list a subset of classes sorted in resolution usage ratio (ratio of ``choosing 224$\times$224/168$\times$168/112$\times$112" over all policies) in Figure \ref{fig:supple-1}(c). Only less than 1\% of frames are used in 84$\times$84 resolution in our experiments, so we omit ``resolution 84" in Figure \ref{fig:supple-1}(a). On dataset level, we observe that AR-Net skips relatively more frames on Mini-Kinetics compared to ActivityNet and FCVID, indicating that videos in Mini-Kinetics are less motion-informative. Moreover, on the class level, samples with complex procedures (e.g. ``making a sandwich" from ActivityNet in Figure \ref{fig:supple-1}(b)) are using more frames with high resolution, compared to the samples with static objects, scenes (``lightning" from FCVID in Figure \ref{fig:supple-1}(b) and (c)) or scene-related actions (``ballet" or ``building cabinet" in Figure \ref{fig:supple-1}(c)), indicating that our learned decision policy often corresponds to the difficulty in making predictions (i.e., difficult samples require more frames with high resolution).
\vspace{-2em}

\newpage
\section{Additional Qualitative Analysis}
\begin{figure}[!htbp]
    \centering
    \includegraphics[width=.89\linewidth]{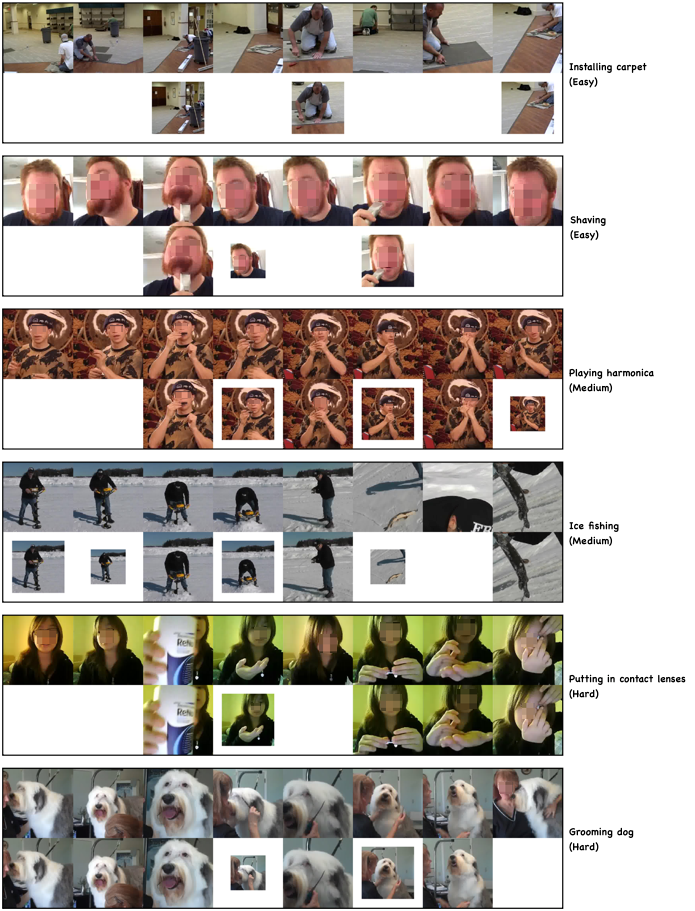}
    \vspace{-1.5em}
    \caption{Qualitative results on ActivityNet dataset. \depict}
    \label{fig:supple-2-actnet}
\end{figure}

\begin{figure}[!htbp]
    \centering
    \includegraphics[width=.999\linewidth]{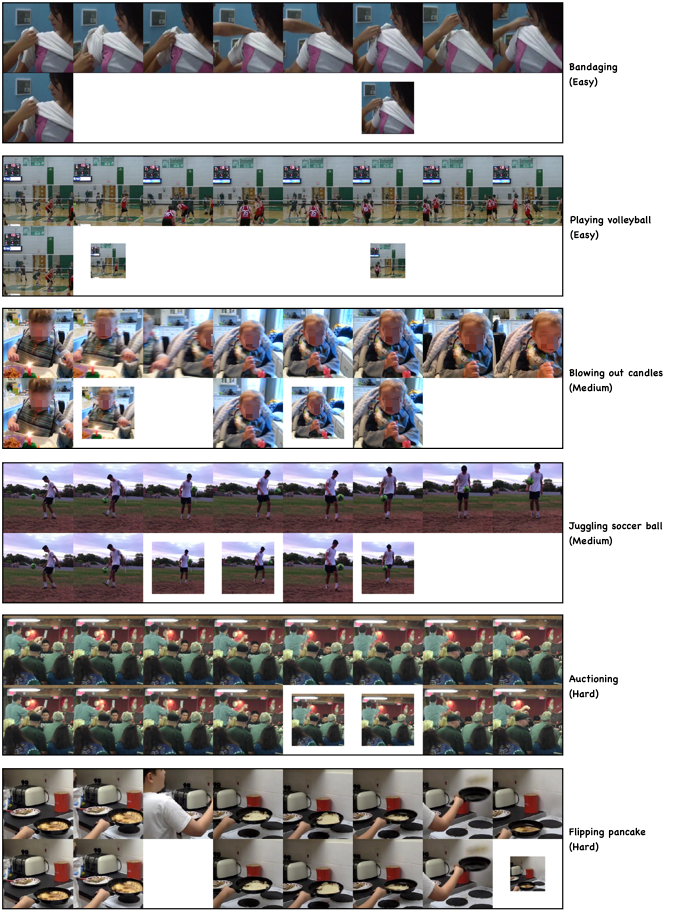} 
    \caption{Qualitative results on Mini-Kinetics dataset. \depict}
    \label{fig:supple-2-minik} 
\end{figure}

\vspace{-0.5em}
Figure \ref{fig:supple-2-actnet} $\sim$ \ref{fig:supple-2-minik} show more qualitative results that AR-Net predicts on ActivityNet, FCVID and Mini-Kinetics. We define ``difficulties" (Easy, Medium and Hard) based on their computation budgets. In general, AR-Net saves the computation greatly for examples that contain clear appearance or actions with less motion.

\vspace{-1.5em}
\vspace{-1.5em}

\newpage
\begin{figure}[!htbp]
    \centering
    \includegraphics[width=.999\linewidth]{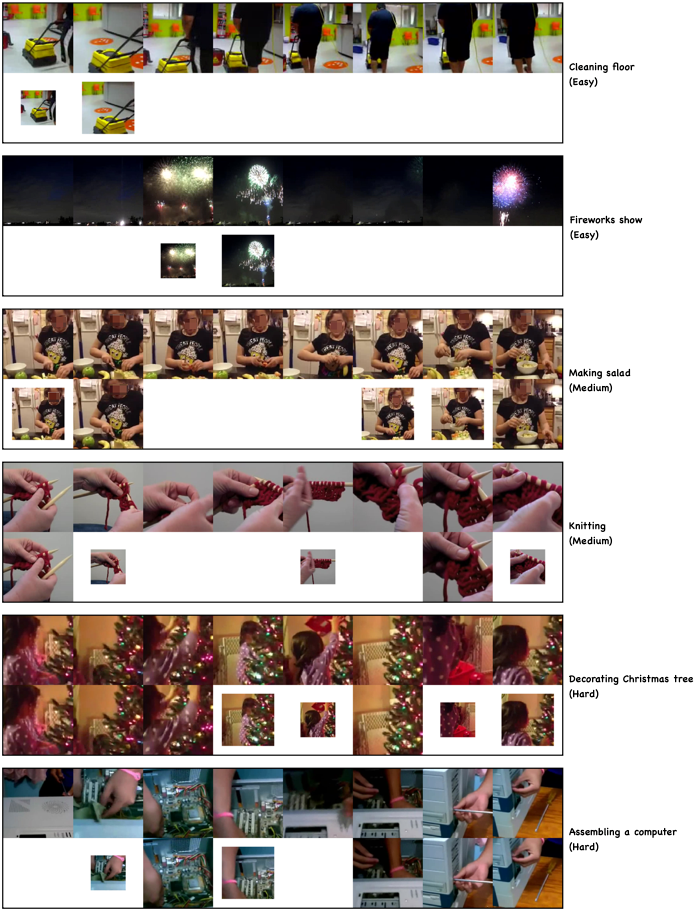} 
    \caption{Qualitative results on FCVID dataset. \depict}
    \label{fig:supple-2-fcvid} 
\end{figure}

\newpage
\clearpage

\section{RL vs Gumbel Softmax in Policy Learning}
We conduct an experiment to compare different policy learning approaches. For the Reinforcement Learning method, we adopt the policy gradient approach and follow the same training procedures and number of epochs used in the Gumbel Softmax experiment. Based on the hyperparameters provided from the previous experiment, we further tune learning rates ($0.001\to 0.002$ in joint-training stage; $0.0005\to 0.001$ in finetuning) to get the best performance for the RL-based method. As shown in Table \ref{table:appendix-2}, Gumbel Softmax approach can achieve a better trade-off in recognition performance (less GFLOPS usage with higher mAP), showing its effectiveness over the RL-based approach.

\begin{table}[!htbp]
\def\arraystretch{1.3}
  \caption{Performances for different learning approaches on ActivityNet-v1.3}
  \begin{center}
  \begin{tabular}{ c | c | c | c } 
 \hline
  
Approach & mAP & GFLOPS/f & GFLOPS/v\\
\hline
Policy Gradient & 72.4 & 3.17 & 50.69\\
Gumbel Softmax & \textbf{73.8} & \textbf{2.09} & \textbf{33.47} \\
\hline
\end{tabular}
\end{center} 
\label{table:appendix-2} 
\end{table}
\nottoggle{Arxiv}{
\section*{Acknowledgement}
This work is supported by IARPA via DOI/IBC contract number D17PC00341. The U.S. Government is authorized to reproduce and distribute reprints for Governmental purposes notwithstanding any copyright annotation thereon. This work is partly supported by the MIT-IBM Watson AI Lab.

\parclaim{Disclaimer} The views and conclusions contained herein are those of the authors and should not be interpreted as necessarily representing the official policies or endorsements, either expressed or implied, of IARPA, DOI/IBC, or the U.S. Government.
}{}

\end{document}